\pdfoutput=1

\documentclass[11pt]{article}

\usepackage[final]{acl}

\usepackage{times}
\usepackage{latexsym}

\usepackage[T1]{fontenc}

\usepackage[utf8]{inputenc}

\usepackage{microtype}

\usepackage{inconsolata}

\usepackage{colortbl} 
\usepackage{xcolor} 
\usepackage{graphicx}
\usepackage{multirow}
\usepackage{multicol}
\usepackage{booktabs}
\usepackage{amssymb}
\usepackage{amsmath}
\usepackage{xspace}
\usepackage{enumitem}
\usepackage{amssymb}
\usepackage{makecell}
\usepackage{amsmath}
\usepackage{xcolor} 

\newcommand{\paratitle}[1]{\vspace{1.5ex}\noindent\textbf{#1}}
\newcommand{\ie}{\emph{i.e.,}\xspace}

\newcommand{\eg}{\emph{e.g.,}\xspace}

\newcommand{\etc}{\emph{etc}}
\newcommand{\ignore}[1]{}

\title{CAFE: Retrieval Head-based Coarse-to-Fine Information Seeking to Enhance Multi-Document QA Capability}

\author{
 \textbf{Han Peng\textsuperscript{1*}},
 \textbf{Jinhao Jiang\textsuperscript{1*}},
 \textbf{Zican Dong\textsuperscript{1*}},
 \textbf{Wayne Xin Zhao\textsuperscript{1\dag}},
 \textbf{Lei Fang\textsuperscript{2}},
\\
 \textsuperscript{1}Gaoling School of Artificial Intelligence, Renmin University of China.
\\
 \textsuperscript{2}DataCanvas Alaya NeW.
\\
   \texttt{\{panospeng, jiangjinhao, dongzican\}@ruc.edu.cn}
\\
   \texttt{batmanfly@gmail.com}
}

\begin{document}
\maketitle
\renewcommand{\thefootnote}{\fnsymbol{footnote}} 
\footnotetext[1]{Equal Contribution.} 
\footnotetext[2]{Corresponding author.}  

\renewcommand{\thefootnote}{\arabic{footnote}}
\begin{abstract}
Advancements in Large Language Models (LLMs) have extended their input context length, yet they still struggle with retrieval and reasoning in long-context inputs. Existing methods propose to utilize the prompt strategy and retrieval head to alleviate this limitation. However, they still face challenges in balancing retrieval precision and recall, impacting their efficacy in answering questions.
To address this, we introduce \textbf{CAFE}, a two-stage coarse-to-fine method to enhance multi-document question-answering capacities. By gradually eliminating the negative impacts of background and distracting documents, CAFE makes the responses more reliant on the evidence documents. Initially, a coarse-grained filtering method leverages retrieval heads to identify and rank relevant documents. Then, a fine-grained steering method guides attention to the most relevant content. 
Experiments across benchmarks show CAFE outperforms baselines, achieving up to 22.1\% and 13.7\% SubEM improvement over SFT and RAG methods on the Mistral model, respectively.
\end{abstract}

\section{Introduction}

\begin{figure}[t]
  \centering
  \includegraphics[width=0.9\columnwidth]{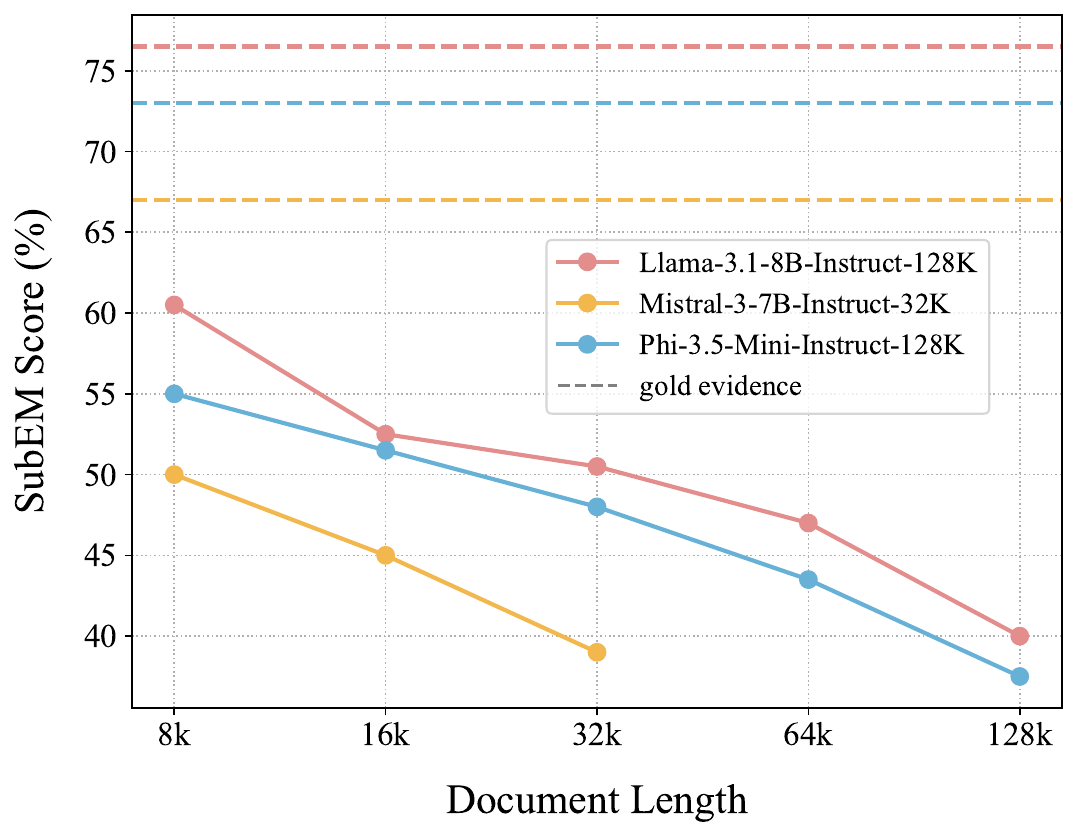}
  \small
  \caption{LLMs' performance on HotpotQA varies with the number of input documents. Solid lines represent performance with the gold document, while dashed lines show performance as more documents are added.}
  \label{fig:exp-gold-other}
\end{figure}

Researchers have undertaken various efforts to extend the context length of Large Language Models (LLMs), ranging from advancements in model architectures~\cite{yen-acl-2024-parallel, munkhdalai-arxiv-2024-infiniattn} to optimizations in training methods~\cite{fu-icml-2024-dataengineering, an-nips-2024-utilize, xiong-naacl-2024-effective, dong-arxiv-2025-longred}. These developments have enabled some recently introduced LLMs to support relatively long context inputs (\ie 128K context length for LLaMA-3.1~\cite{Dubey-arxiv-2023-llama3} and Qwen-2.5~\cite{yang-arxiv-2024-qwen}, and even 10M context length for Gemini~\cite{reid-arxiv-2024-gemini}). However, recent studies indicate that LLMs exhibit limitations in retrieval and reasoning capability when processing the long context input~\cite{liu-tacl-2024-lostmiddle, lee-arxiv-2024-can, wang-emnlp-2024-leave, li-acl-2024-multihopreason}, which poses significant challenges for their effective application in downstream tasks, including book summarization~\cite{bai-acl-2024-longbench}, multi-document question answering~\cite{zhang-arxiv-2024-infty}, and code repository understanding~\cite{bai-arxiv-2024-longbenchv2}.

To investigate performance bottlenecks, we conduct preliminary experiments on long-context retrieval and reasoning, focusing on multi-document question answering, a representative task recently studied~\cite{zhu-acl-2024-fanoutqa, hsieh-arxiv-2024-ruler}. Our findings reveal that the performance of LLMs significantly declines as the number of additional input documents increases, compared to using only the necessary documents (\ie, gold documents), as shown in Figure~\ref{fig:exp-gold-other}. This prompts a natural question: \textit{how can we mitigate the impact of these additional documents in long-context input?}

To address this question, existing studies~\cite{agrawal-emnlp-2024-r&r, zhang-arxiv-2024-autopasta} primarily employ external prompt strategies to guide models in first extracting key information related to a question from the long input and then utilizing them along with the original context, to answer the question. However, its effectiveness is constrained by the LLMs' instruction-following capabilities~\cite{xu-iclr-2024-retrievalmeetslclm}, which is amplified in long inputs. To address this issue, subsequent research~\cite{yifu-arxiv-2025-eliciting} analyses the model's internal attention mechanism, utilizing the retrieval head to identify key information. As shown in Table~\ref{tab:recall-icr-attention}, our experiments also indicate that the attention head outperforms the external prompt strategy. Nonetheless, these methods face challenges in balancing retrieval precision and recall, impacting their efficacy in answering questions. Specifically, enhancing recall introduces more irrelevant information during the model's reasoning process, while increasing retrieval precision reduces recall.

To address these challenges, we first rethink that humans typically do not search across all documents; instead, they would gradually \emph{identify} the relevant ones step-by-step and then \emph{reason} over them. For example, individuals first filter out irrelevant documents to create a manageable candidate set, then further analyze and utilize the most relevant documents to answer the question. Inspired by this, we aim to design a strategy that guides LLMs to gradually filter out irrelevant documents from the long-context input, and then further enhance their ability to utilize the remaining relevant documents to answer questions. 

According to the above motivation, we propose \textbf{CAFE}, a novel two-stage coarse-to-fine information-seeking method to enhance the multi-document question-answering capabilities of LLMs. Its core idea lies in a fine-grained utilization of retrieval heads based on specific contexts in multiple stages. 
Specifically, before information-seeking, we pre-locate the retrieval heads for the two stages respectively on the validation set. Then, in the first stage, we implement a coarse-grained filtering approach to filter out background documents. We identify relevant documents assigned with high attention scores in each pre-detected retrieval head and further rerank these documents according to the summed scores from all retrieval heads. In the second stage, we guide the model using a fine-grained steering approach. We utilize another set of retrieval heads to further select relevant documents from these reranked documents, and employ attention steering on the most relevant content to answer the final questions. In this way, we can guide the LLMs to gradually search for evidence documents in the long context input and utilize them to better answer the questions. Additionally, the whole method is training-free and applicable to a wide range of downstream tasks.

We conducted extensive experiments to evaluate the proposed CAFE method using various LLMs. The results demonstrate that our method consistently outperforms existing strong baselines across five benchmarks and three LLMs (\eg achieving an 11.4\% relative performance improvement compared to the supervised fine-tuning method).



\section{Related Work}

\begin{table*}[t]
  \centering
  \small
  \begin{tabular}{lccccc}
    \toprule
    \textbf{Method} & \textbf{HotpotQA-8k} & \textbf{HotpotQA-16k} & \textbf{HotpotQA-32k} & \textbf{SQuAD} & \textbf{Musique} \\
    \midrule
    ICR & 0.64 & 0.51 & 0.38 & 0.91 & 0.58 \\
    Attention-Based & \textbf{0.80} & \textbf{0.78} & \textbf{0.77} & \textbf{0.94} & \textbf{0.79} \\
    \bottomrule
  \end{tabular}
  \caption{Recall scores for different evidence selection strategies across various datasets using Llama-3.1-8B-Instruct.}
  \label{tab:recall-icr-attention}
\end{table*}

\label{sec:related_work}
\paratitle{Long-Context Utilization in Language Models.}
Although extended the context length of LLMs successfully~\cite{Dubey-arxiv-2023-llama3,yang-arxiv-2024-qwen,dong-nips-2025-exploring}, they still face significant challenges (\eg long-term decay~\cite{chen-arxiv-2024-hope} and lost-in-the-middle~\cite{liu-tacl-2024-lostmiddle}) in utilizing long contexts effectively for complex tasks.
To enhance the long-context utilization capacities, attention-based methods leverage the property of attention heads and positional encodings, enlarging the attention scores of the key tokens over the long inputs~\cite{wu-arxiv-2024-retrieval, pradipta-arxiv-2024-decore}. Different from previous methods, our work employs a training-free two-stage framework, which identifies relevant documents and guild the response more dependent on these documents.





\paratitle{Retrieval Head in Attention Mechanisms.}
Recent studies have revealed specialized attention heads in LLMs that exhibit retrieval capabilities for locating critical information within long contexts, namely, retrieval heads~\cite{wu-arxiv-2024-retrieval}. In these heads, high attention values will be assigned to the tokens most relevant to the current token in the long inputs, achieving in-context retrieval of previous information. Recently, some work retains the full attention on the retrieval heads and employs KV Cache compression on other heads to accelerate the calculation~\cite{fu-arxiv-2024-moa, tang-arxiv-2024-razorattention,xiao-arxiv-2024-duoattetnion}. Different from them, our method utilizes retrieval heads as a retrieval system to identify evidence documents. 


\paratitle{Retrieval-Augmented Generation.}
Retrieval-Augmented Generation (RAG) has been widely adopted to address various NLP tasks.
For multi-document question-answering tasks, traditional RAG methods utilize external dense or sparse retrieval models to compute the similarity of documents with the question~\cite{robertson-ftir-2009-bm25, karpukhin-emnlp-2020-dense}. Then, relevant documents are retrieved as the input for models.
Beyond leveraging external models to retrieve documents, several in-context retrieval methods have been proposed~\cite{agrawal-emnlp-2024-r&r,li-arxiv-2024-alr2}. These methods prompt the models to select the indices of relevant documents. 
Unlike existing RAG approaches, our work leverages the model’s inherent retrieval capabilities to perform a coarse-to-fine location of evidence documents, effectively enhancing its retrieval and reasoning abilities.
\section{Empirical Study}

\label{sec:empirical_study}
In this section, we conduct empirical studies to analyze how to improve the retrieval and reasoning capabilities of LLMs for multi-document question answering from two aspects, \ie evidence selection and attention intervention.



\subsection{Evidence Selection}
\label{sec:evidence_selection}

In this study, we examine document retrieval in multi-document question answering using two primary methods: in-context retrieval and attention-based retrieval. The in-context retrieval method prompts LLMs to directly select the top-\(k\) documents most relevant to the given question, while attention-based retrieval method utilizes the attention distribution over each document to perform selection. Subsequently, the selected documents are employed as condensed input in the question-answering prompt. The results, presented in Table~\ref{tab:recall-icr-attention}, demonstrate that the attention-based approach significantly outperforms the in-context retrieval method, particularly as document length increases (e.g., recall score decreases by 50.6\% on HotpotQA-32K). More experimental results can be found in Appendix~\ref{app:evidence}.

\begin{table}[t]
  \centering
  \small
  \begin{tabular}{lc}
    \toprule
    \textbf{Mask Mode} & \textbf{SubEM} \\
    \midrule
    No Mask & \textbf{60.5} \\
    $\text{Evidence}_{2}$ $\rightarrow$ $\text{Evidence}_{1}$ & 60.0 \\
    Question $\rightarrow$ $\text{Evidence}_{1}$ & 48.0 \\
    Question $\rightarrow$ $\text{Evidence}_{2}$ & 42.0 \\
    Question $\rightarrow$ $\text{Evidence}_{1}$, $\text{Evidence}_{2}$ & 29.5 \\
    Question $\rightarrow$ $\text{One Irrelevant Document}$ & 60.0 \\
    Question $\rightarrow$ $\text{Two Irrelevant Documents}$ & 59.5 \\
    
    \bottomrule
  \end{tabular}
  \caption{SubEM scores on HotpotQA-8K with various masking strategies using Llama-3.1-8B-Instruct, where \(\text{Evidence}_{1}\) and \(\text{Evidence}_{2}\) refer to the first and second gold documents in the context.}
  \label{tab:attention-flow}
\end{table}

\subsection{Attention Intervention}\label{sec:attention_intervention}
Through the above experiments, we find that using attention heads can effectively help the model to retrieve the most relevant documents from long inputs. In this section, we further explore how to facilitate the model's utilization of multiple retrieved documents. Specifically, we select HotpotQA test samples with input lengths within 8K tokens. Based on the attention interaction, we mask the attention between the two gold documents, as well as mask the attention of the questions to the two gold documents, respectively. We show the results in Table~\ref{tab:attention-flow}. First, masking the attention between the two gold documents has negligible impact on performance compared to the unmasked condition (60.0\% for Evidence$_2$ $\rightarrow$ Evidence$_1$ vs. 60.5\% for No Mask). This suggests that during multi-hop question answering, the LLM does not engage in implicit reasoning while encoding long inputs, aligning with observations from recent studies~\cite{yu-emnlp-2024-incontext}. 
Implicit reasoning in our work denotes the model’s capacity to integrate information across documents during the prefilling phase, before the question is posed. By using attention interactions, the model embeds pertinent content from earlier documents into later ones, allowing it to retrieve the answer directly from the final document without revisiting prior texts.

\begin{figure*}[t]
  \includegraphics[width=\linewidth]{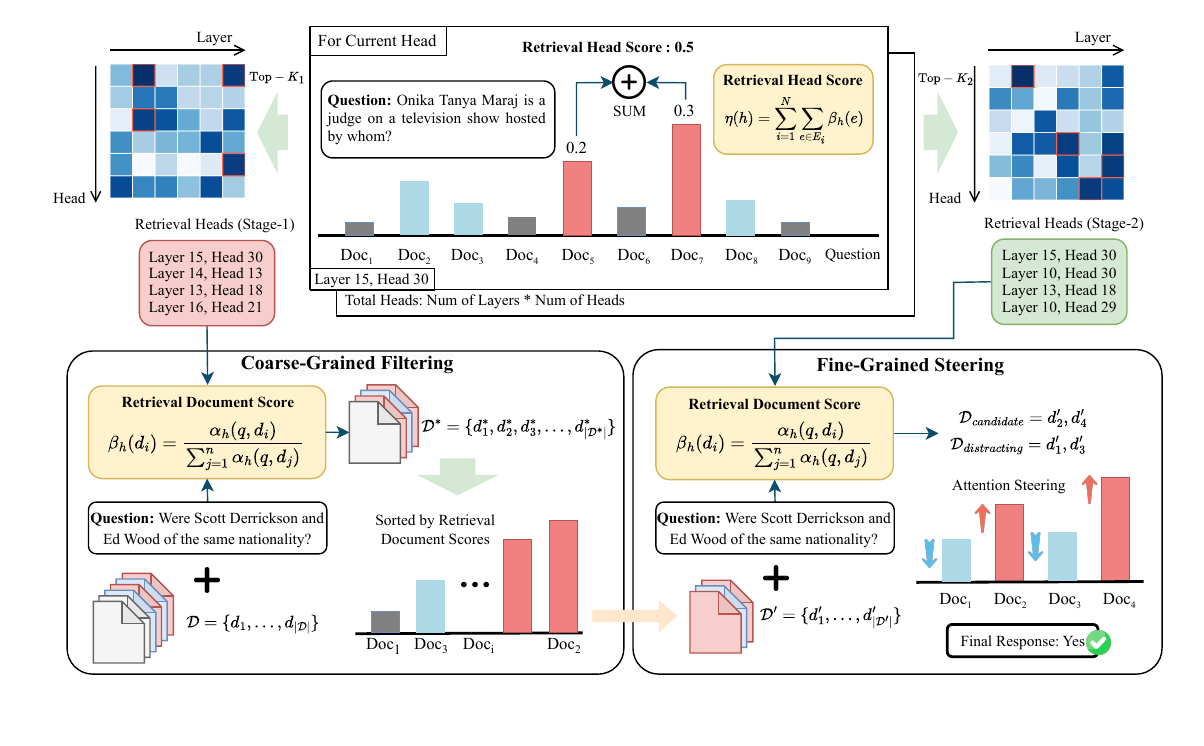}
  \caption {Overall framework of our proposed CAFE approach. The red, blue, and yellow bar charts represent the gold, distracting, and background documents, respectively.}
\label{fig:method}
\end{figure*}

Moreover, when applying attention masks to question-irrelevant documents, we observed minimal performance impact. Additionally, masking the attention from the question to any gold document results in a significant performance drops. When all gold documents are masked simultaneously, the SubEM score even decreases to a level similar to that observed when no document is provided (29.5\% for Question $\rightarrow$ $\text{Evidence}_{1}$, $\text{Evidence}_{2}$ vs. 23.0\% for Closed Book). This demonstrates that the directly attention from the question to the gold documents plays a critical role in the LLM's overall performance. 

Building on the preceding experiments, attention heads can be employed to assist the LLM in retrieving pertinent documents from long inputs. In addition, By adjusting the attention directed towards these documents, the model's ability to utilize them in answering questions is enhanced. These principles will guide the design of our method.
\section{Method}
\label{sec:method}

\subsection{Overall Framework}
\label{sec:method_overview}
In multi-document question-answering tasks, there are three categories of documents, \ie gold evidence documents that contain information supporting answering the questions, distracting documents that impede the model's ability to generate faithful answers, and background documents that contain irrelevant information. Among them, The latter two categories of documents increase the text length and introduce noise, which further weaken the model's capabilities. Thus, we propose \textbf{CAFE}, a coarse-to-fine two-stage framework to enhance the long-context question-answering capacities by gradually eliminating the negative impacts of background and distracting documents. 
In our framework, we identify retrieval heads to locate background and distracting documents from the input set. Given that the two stages rely on different context information, we use different retrieval heads for each. We first apply coarse-grained filtering to remove background documents, then use fine-grained attention steering to reduce the influence of distracting ones. This two-stage process helps the model focus on gold evidence. The overall illustration is shown in Figure~\ref{fig:method}.


\subsection{Retrieval Head Detection}
\label{sec:retrieval_head}
In Section~\ref{sec:empirical_study}, we observe that the attention scores can effectively identify evidence documents. Additionally, there exist some heads where the attention scores of tokens in the question usually focus on tokens within the relevant content, namely, retrieval heads~\cite{wu-arxiv-2024-retrieval}. Leveraging the properties, we first identify retrieval heads that can be further employed to seek the relevant documents.


\paratitle{Retrieval Document Scores.} 
Based on the analysis in Section~\ref{sec:attention_intervention}, we focus on the attention distribution from the question to the contextual documents. 
Therefore, 
we first compute the \textbf{retrieval document score} $\beta_h(d_i)$ by analyzing attention weight distributions between the question $q$ and each document $d_i$:
\begin{equation}
\beta_h(d_i) = \frac{\alpha_h(q, d_i)}{\sum_{j=1}^{n} \alpha_h(q, d_j)},
\end{equation}
where $ \alpha_h(q, d_i) $ represents the attention weight between the query $ q $ and document $ d_i $ for attention head $ h $, and $ n $ is the total number of documents in the current sample.

\paratitle{Top-$K$ Retrieval Heads Selection.}
\label{sec:retrievalhead_selection}
To effectively identify retrieval heads, we select $N$ samples from the validation set and calculate a \textbf{retrieval head score} for each attention head $h$ based on the evidence documents' retrieval document scores on these validation samples:
\begin{equation}
\eta(h) = \sum_{i=1}^{N} \sum_{e \in E_i} \beta_h(e),
\end{equation}
where $ E_i $ is the set of evidence documents for the $i$-th sample. Subsequently, we select the {Top-$K$} attention heads $\mathcal{H}_{\text{ret}}$ with the highest {retrieval head scores} from the heads from all layers $\mathcal H$ as the retrieval heads. 
\begin{equation}
    \mathcal H_{ret} = \text {Top-}K (\eta(h)), ~~ h \in \mathcal H.
\end{equation}
Notably, during the coarse-grained filtering and fine-grained steering stages, we employ different validation sets and select different retrieval heads according to the properties of the two stages. The distinction between the two types of retrieval heads is detailed in Appendix~\ref{app:diff_retrieval_heads}.

\subsection{Coarse-Grained Filtering for Background Documents}
\label{sec:coarse-grained_filtering}

In Figure~\ref{fig:exp-gold-other}, we observe that a large amount of background documents leads to significant performance degradation. Thus, we introduce a coarse-grained filtering stage to filter background documents and obtain a condensed input. Specifically, this stage consists of two steps: background document filtering and locality-based re-ranking. 


\paratitle{Background Documents Filtering.} To identify background documents, we first compute the retrieval document scores of each document on selected retrieval heads $\mathcal H_{ret}$. For each head $h$, we select Top-$M_1$ documents based on the retrieval document scores $\beta_h(d)$ from all documents $\mathcal D$ and consider them as relevant documents. Then, we perform a union operation on these documents to obtain the relevant document set $\mathcal D^*$ and drop the other documents.
\begin{equation}
\mathcal{D}^* = \bigcup_{h \in \mathcal{H}_{\text{ret}}} \text{Top-}{M_1} (\beta_h(d)), ~~ d \in \mathcal D.
\end{equation}

\paratitle{Locality-Based Re-Ranking.}
When processing long context, LLMs usually demonstrate the property of locality and lost-in-the-middle~\cite{liu-tacl-2024-lostmiddle,su-neurocomputing-2024-roformer}. This means when critical information for answering the question is located at the end of the long document, the model often performs better. Thus, after obtaining the filtered set of documents $\mathcal{D}^*$, we apply a \textbf{locality-based re-ranking} mechanism to rank these documents. 
For the filtered candidate document set $\mathcal{D}^*$, we compute the \textbf{document relevance score} $\gamma_{h}(d)$ for each document as the sum of retrieval document scores of all retrieval heads:
\begin{equation}
\gamma_{h}(d) = \sum_{h \in \mathcal{H}_{\text{ret}}} \beta_h(d), ~~d \in \mathcal{D}^*.
\end{equation}
Subsequently, documents with higher document relevance scores are positioned later in the sequence, ensuring that more attention will be focused on the documents that are more likely to contain critical evidence during the generation of responses. 
Finally, we obtain the filtered and reranked document sequence $\mathcal D'$ as the input of next stage:
\begin{align}
    \mathcal D' = \{d'_1, \dots, d'_{\lvert D^*\rvert}\}, \forall i < j,  \gamma_h(d_i) \leq \gamma_h(d_j).
\end{align}

\subsection{Fine-Grained Steering for Distracting Text}
\label{sec:fine-grained_steering}

\begin{table*}[h]
    \centering
    \resizebox{\textwidth}{!}{ 
    \begin{tabular}{llcccccc|cccccc}
        \toprule
        \multirow{2}{*}{\textbf{LCLM}} & \multirow{2}{*}{\textbf{Baseline}} & \multicolumn{2}{c}{\textbf{SQuAD}} & \multicolumn{2}{c}{\textbf{MuSiQue}} & \multicolumn{2}{c|}{\textbf{HotpotQA}} & \multicolumn{2}{c}{\textbf{HotpotQA-16K}} & \multicolumn{2}{c}{\textbf{HotpotQA-32K}}\\

        \cmidrule(lr){3-4} \cmidrule(lr){5-6} \cmidrule(lr){7-8} \cmidrule(lr){9-10} \cmidrule(lr){11-12}
        & & \textbf{SubEM} & \textbf{F1} & \textbf{SubEM} & \textbf{F1} & \textbf{SubEM} & \textbf{F1} & \textbf{SubEM} & \textbf{F1} & \textbf{SubEM} & \textbf{F1} \\
        \midrule
        \multirow{8}{*}{Llama-3.1-8B}
        & Oracle RAG   & 92.5 & 86.4 & 39.0 & 39.3 & 76.5 & 76.8 & 76.5 & 76.8 & 76.5 & 76.8 \\
        \cmidrule(lr){2-12}
        & Directly Answering  & 71.0 & 66.6 & 30.5 & 33.2 & 60.5 & 62.5 & 53.0 & 60.1 & 53.5 & 58.1 \\
        & In-Context Retrieval  & 73.5 & 65.1 & 28.0 & 29.2 & 59.0 & 58.6 & 51.5 & 51.8 & 42.5 & 42.2 \\
        & Vanilla RAG  & 84.5 & 76.6 & 28.0 & 29.7 & 64.0 & 64.8 & 63.0 & 63.6 & 61.5 & 62.4 \\
        & SFT  & 69.0 & 70.1 & 33.5 & \textbf{38.9} & 63.0 & \underline{69.8} & 62.5 & 68.0 & 61.5 & \underline{67.4} \\
        \cmidrule(lr){2-12}
        & CAFE (w/o FGS) & \underline{89.5} & \underline{80.7} & \underline{36.0} & 35.5 & \underline{68.5} & 69.0 & \underline{66.0} & \underline{68.3} & \underline{66.0} & 65.2 \\
        & CAFE~(ours) & \textbf{89.5} & \textbf{82.6} & \textbf{36.5} & \underline{36.5} & \textbf{70.0} & \textbf{70.4} & \textbf{69.0} & \textbf{69.0} & \textbf{68.5} & \textbf{68.1} \\
        \midrule
        
        \multirow{8}{*}{Mistral-3-7B}
        & Oracle RAG   & 84.0 & 80.1 & 40.5 & 38.9 & 67.0 & 71.3 & 67.0 & 71.3 & 67.0 & 71.3 \\
        \cmidrule(lr){2-12}
        & Directly Answering  & 59.0 & 55.9 & 27.5 & 26.8 & 50.0 & 53.7 & 45.0 & 47.5 & 39.0 & 46.6 \\
        & In-Context Retrieval  & 59.5 & 58.7 & 24.0 & 24.2 & 49.0 & 47.6 & 37.5 & 38.2 & 29.5 & 30.3 \\
        & Vanilla RAG  & 69.5 & 69.2 & 27.5 & 26.2 & 53.5 & 55.9 & 53.5 & 55.4 & 51.0 & 54.7 \\
        & SFT  & 60.0 & 60.1 & \underline{30.5} & \textbf{33.1} & 57.5 & 61.9 & 52.5 & 56.7 & 47.5 & 53.6 \\
        \cmidrule(lr){2-12}
        & CAFE (w/o FGS) & \underline{78.0} & \underline{73.6} & 30.0 & 27.9 & \underline{60.0} & \underline{64.0} & \underline{60.5} & \underline{60.0} & \underline{53.0} & \underline{56.5} \\
        & CAFE~(ours) & \textbf{78.5} & \textbf{75.2} & \textbf{31.0} & \underline{29.9} & \textbf{61.5} & \textbf{65.2} & \textbf{60.5} & \textbf{61.7} & \textbf{58.0} & \textbf{61.7} \\
        \midrule
        
        \multirow{8}{*}{Phi-3.5-Mini}
        & Oracle RAG   & 85.0 & 80.0 & 35.0 & 38.1 & 73.0 & 75.8 & 73.0 & 75.8 & 73.0 & 75.8 \\
        \cmidrule(lr){2-12}
        & Directly Answering  & 63.5 & 58.8 & 24.5 & 27.5 & 55.0 & 55.5 & 51.5 & 52.5 & 48.0 & 48.3 \\
        & In-Context Retrieval  & 65.5 & 66.4 & 22.5 & 23.7 & 49.5 & 49.5 & 38.0 & 39.5 & 31.0 & 34.4 \\
        & Vanilla RAG  & 76.0 & 72.5 & 25.5 & 26.1 & 58.5 & 60.2 & 56.0 & 58.8 & 55.0 & 58.7 \\
        & SFT  & 64.5 & 65.1 & \textbf{34.5} & \textbf{40.9} & 60.5 & \textbf{71.8} & 61.0 & \textbf{71.8} & 58.0 & \textbf{67.3} \\
        \cmidrule(lr){2-12}
        & CAFE (w/o FGS) & \underline{82.0} & \underline{74.9} & 28.5 & 28.8 & \underline{65.0} & 67.8 & \underline{64.5} & 62.6 & \underline{60.0} & 58.9 \\
        & CAFE~(ours) & \textbf{84.5} & \textbf{75.8} & \underline{30.0} & \underline{31.9} & \textbf{66.5} & \underline{68.0} & \textbf{66.5} & \underline{64.8} & \textbf{61.5} & \underline{60.1} \\
        \bottomrule
    \end{tabular}
    }
    \caption{Evaluation results on three long-document question answering tasks. They are representative of single-hop and multi-hop question-answering tasks. ``CAFE (w/o FGS)'' means that we only perform the fist stage without the fine-grained steering for distracting text stage. The \textbf{bold} and \underline{underline} fonts denote the best and second best results in each dataset. Notably, all models in the table are the instruct versions.}
    \label{tab:results}
\end{table*}

After the first stage of filtering background-irrelevant documents, though the remaining documents usually contain question-relevant information, there may still be distracting documents that are not useful for answering the question. Thus, in the fine-grained steering stage, we further identify these distracting documents and steering the attention scores on them to reduce the influence of these documents on the generation of responses.

\paratitle{Iterative Distracting Document Identification.}
Similar to the coarse-grained filtering stage, to effectively identify and weaken the impact of these distracting documents, we perform document identification by computing retrieval document scores using another set of retrieval heads $\mathcal{H}_{\text{ret}}'$:

\begin{equation}
\mathcal{D}_\text{cand} = \bigcup_{h \in \mathcal{H}_{\text{ret}}'} \text{Top-}M_2 (\beta_h( d)),~ d \in \mathcal D'
\end{equation}
By identifying documents with high retrieval document scores, we ultimately derive a candidate set of evidence documents $\mathcal{D}_\text{cand}$. Each document in the candidate set is considered the golden evidence while other documents are considered as distracting documents during the following process of attention steering. 


\paratitle{Inference-Time Attention Steering.}
After the initial filtering stage, the number of remaining documents is significantly reduced. In this stage, directly removing detected distractors may result in lower recall of evidence documents. Thus, instead of only keeping the candidate set, 
we adopt post-hoc attention steering ~\cite{zhang-iclr-2024-pasta}, an inference-only technique that reweights attention scores to guide the model's focus toward user-specified input spans. 
Specifically, given the candidate gold evidence set $\mathcal{D}_\text{cand}$, our method emphasizes specific tokens by adding a constant attention bias $\mathbf B^h$ to the attention scores on tokens within these documents across all attention heads.
\begin{align}
    \tilde{\mathbf A}^h &= \mathrm{Softmax}(({\mathbf Q^h}^\intercal {\mathbf K^h}+\mathbf B^h)/\sqrt d ),\\
    B^h_{ij}& = \begin{cases}
    \delta & \text{if } i \in q  \text{ and } j \in \mathcal{C}_\text{cand} \\
    0 & \text{otherwise}
    \end{cases},
\end{align}
where $\delta$ is a positive constant that controls the degree of attention adjustment. After applying $\text{Softmax}(\cdot)$, the attention scores of tokens in $\mathcal{D}_\text{cand}$ are enlarged while the attention scores of other tokens are reduced.
This dynamic reweighting mechanism effectively enhances the model's attention toward tokens in $\mathcal{D}_\text{cand}$, ensuring the responses are more dependent on the critical evidence. 


\section{Experiments}
\label{sec:experiments}

\subsection{Experimental Setup}
\label{sec:experimental_setup}

\paratitle{Datasets.} We evaluate the long-context performance of our approach and baseline methods using three question-answering datasets: SQuAD \citep{rajpurkar-emnlp-2016-squad}, HotpotQA \citep{yang-emnlp-2018-hotpotqa}, and MusiQue \citep{trivedi-tacl-2022-musique}. These datasets are collected from the RULER \citep{hsieh-arxiv-2024-ruler} and LongBench \citep{bai-acl-2024-longbench} benchmarks. Additionally, we experiment with three versions of HotpotQA that vary in context length to analyze how model performance changes with text length. To ensure consistency across all baselines and our approach, we randomly select 200 samples from each dataset to form the final test set. All experiments are conducted using the same test sets.

\paratitle{Baselines and Metrics.} For evaluation, we use Substring Exact Match (SubEM) and F1 scores following existing work~\cite{li-arxiv-2024-selfimprove}. SubEM measures whether the gold answer appears as a substring in the predictions, while the F1 score evaluates the token-level overlap between predictions and references. For compared baselines, we select five types of methods, including \textit{Directly Answering}, \textit{In-Context Retrieval}, \textit{Oracle RAG}, \textit{Vanilla RAG}, and \textit{Supervised Fine-tuning}. We present the detailed description in Appendix~\ref{app:baselines}. 

\paratitle{Implementation Details.}
We conduct our experiments on three open-source models: Llama-3.1-8B-Instruct, Mistral-3-7B-Instruct, and Phi-3.5-Mini-Instruct. For coarse-grained filtering for background documents, we set the $\text{Top-}{M_1}$ to 4 and $\text{Top-}{K_1}$ to 4. For fine-grained steering for distracting text, we set the $\text{Top-}{M_2}$ to 2 and $\text{Top-}{K_2}$ to \{1,2,3,4\} and we set $\delta = \log10$. As for the SFT configuration, training is conducted with a batch size of 64 and a learning rate of $1 \times 10^{-5}$. We set the number of training rounds to 1, as multiple rounds resulted in overfitting with the limited data available.

\subsection{Main Results}
\label{sec:main_results}
Table~\ref{tab:results} shows the results of our methods and other baselines across three representative long context question-answering datasets.

Firstly, our method achieves significantly better multi-document question-answering performances than other baselines. Across all three datasets, our method consistently outperforms training-free approaches and even surpasses the SFT method in most settings. On single-hop SQuAD, our method can achieve performances nearly the performance ceiling introduced by Oracle RAG. On more complex multi-hop question-answering tasks, our method can still achieve a significant performance improvement (\eg approximately $19.9\%$ of SubEM scores on the HotpotQA dataset compared to the naive directly answering method).


Secondly, the two stages of our method work together to prompt performance improvements. Compared with in-context retrieval and vanilla RAG which retrieve relevant documents via prompting techniques or external models, only employing the coarse-grained filtering stage can greatly boost the performance, indicating that leveraging the inner retrieval heads can more effectively identify relevant documents. Additionally, introducing fine-grained attention steering can further boost long-context question-answering capacities, which demonstrates the necessity of introducing a fine-grained elimination of the negative impacts of distracting documents on multi-document question answering.


Finally, our method exhibits less performance drop with longer input lengths. On the HotpotQA dataset, we assess the performances across different input lengths. Our method can preserve performance to a greater extent when dealing with longer texts (\eg decreases $1.4\%$ and $2.1\%$ SubEM scores for Llama-3.1-8B on 16K and 32K). Instead, the performances drop sharply with the length increasing with other methods, especially in-context retrieval (\eg decreases $12.7\%$ and $28.0\%$ SubEM scores for Llama-3.1-8B on 16K and 32K). This indicates that our method can effectively identify the critical documents in the long input, scarcely affected by the increased number of documents.

\begin{table}[t]
    \centering
    \small
    \begin{tabular}{lccc}
        \toprule
        \textbf{Method} & \textbf{Llama} & \textbf{Mistral} & \textbf{Phi}\\
        \midrule
        \rowcolor{gray!20} CAFE & 70.0 & 61.5 & 66.5 \\ 
        \midrule 
        \text{w/o CGF} & 62.5 & 52.0 & 55.0 \\ 
        \text{w/o FGS} & 68.5 & 60.0 & 65.0 \\
        \text{w/o Re-Ranking} & 68.0 & 59.5 & 65.5 \\
        \bottomrule
    \end{tabular}
    \caption{Ablation study on HotpotQA.} 
    \label{tab:hotpotqa_ablation}
\end{table}

\begin{table}[t]
    \centering
    \small
    \begin{tabular}{lccc}
        \toprule
        \textbf{Granularity} & \textbf{Recall} & \textbf{SubEM} & \textbf{F1}\\
        \midrule
        \text{w/o Steering} & - &  68.5 & 69.0 \\
        \midrule
        \text{Sentence-Level} & 0.89 &  65.5 & 67.8 \\
        \text{Document-Level} & 0.93 & 70.0 & 70.4 \\
        \bottomrule
    \end{tabular}
    \caption{Results with different steering granularities.}
    \label{tab:hotpotqa_granularity}
\end{table}

\subsection{Further Analysis}
\label{sec:further_analysis}

\begin{table*}[t]
  \centering
  \small
  \setlength{\tabcolsep}{4pt}
  \begin{tabular}{llccccccc}
    \toprule
    \textbf{Model} & \textbf{Method} 
    & \textbf{1} & \textbf{10} & \textbf{20} & \textbf{30} & \textbf{40} & \textbf{50} & \textbf{Rand} \\
    \midrule
    \multirow{2}{*}{LLaMA-3.1-8B-Instruct} 
    & DA
    & 77.5/70.9 & 74.5/67.4 & 73.0/67.6 & 70.5/64.6 & 69.5/64.4 & 73.0/67.6 & 71.0/66.6 \\
    & Ours 
    & \textbf{90.5/82.1} & \textbf{91.0/80.3} & \textbf{89.5/80.9} & \textbf{89.0/80.5} & \textbf{88.5/79.9} & \textbf{89.5/79.2} & \textbf{89.5/82.6} \\
    \midrule
    \multirow{2}{*}{Mistral-3-7B-Instruct} 
    & DA
    & 70.0/58.7 & 59.0/50.6 & 56.5/48.3 & 59.5/51.6 & 58.0/52.9 & 62.0/59.5 & 59.0/55.9 \\
    & Ours 
    & \textbf{79.5/76.0} & \textbf{80.0/74.8} & \textbf{79.0/71.9} & \textbf{78.5/71.1} & \textbf{78.5/71.2} & \textbf{78.0/72.6} & \textbf{78.5/75.2} \\
    \bottomrule
  \end{tabular}
  \caption{Position-wise SubEM/F1 scores on two models. The column headers (1, 10, 20, \etc) indicate the document index where the gold document is inserted. DA denotes Directly Answering.}
  \label{tab:lost_in_middle}
\end{table*}

\paratitle{Ablation Study.}
To assess the effectiveness of our framework, we conduct ablation experiments focusing on key steps within the pipeline. 
(1) \textit{w/o Coarse-Grained Filtering~(CGF)} eliminates the initial coarse-grained filtering of background documents;
(2) \textit{w/o Fine-Grained Steering~(FGS)} omits the fine-grained steering of distracting text, relying solely on documents $\mathcal{D}'$ for inference;
(3) \textit{w/o Locality-Based Re-Ranking} bypasses locality-based re-ranking in the first stage, resulting in the use of filtered documents in a random order.

The results are presented in Table~\ref{tab:hotpotqa_ablation}. All variants show inferior performance compared to the original method, underscoring the effectiveness of each component in our framework. Notably, the absence of Coarse-Grained Filtering (\textit{w/o CGF}) results in a substantial performance decline, highlighting the critical role of first-stage filtering in excluding irrelevant background documents and preventing the dilution of the model's attention. Similarly, the removal of Fine-Grained Steering (\textit{w/o FGS}) leads to decreased performance, indicating that the second stage's attention steering effectively mitigates the impact of distracting documents. Furthermore, the exclusion of Re-Ranking (\textit{w/o Re-Ranking}) results in significant performance degradation, demonstrating the effectiveness of putting the essential information at the end of the input, to facilitate retrieval and reasoning of models.

\begin{figure}[t]
\centering
  \includegraphics[width=0.9\columnwidth]{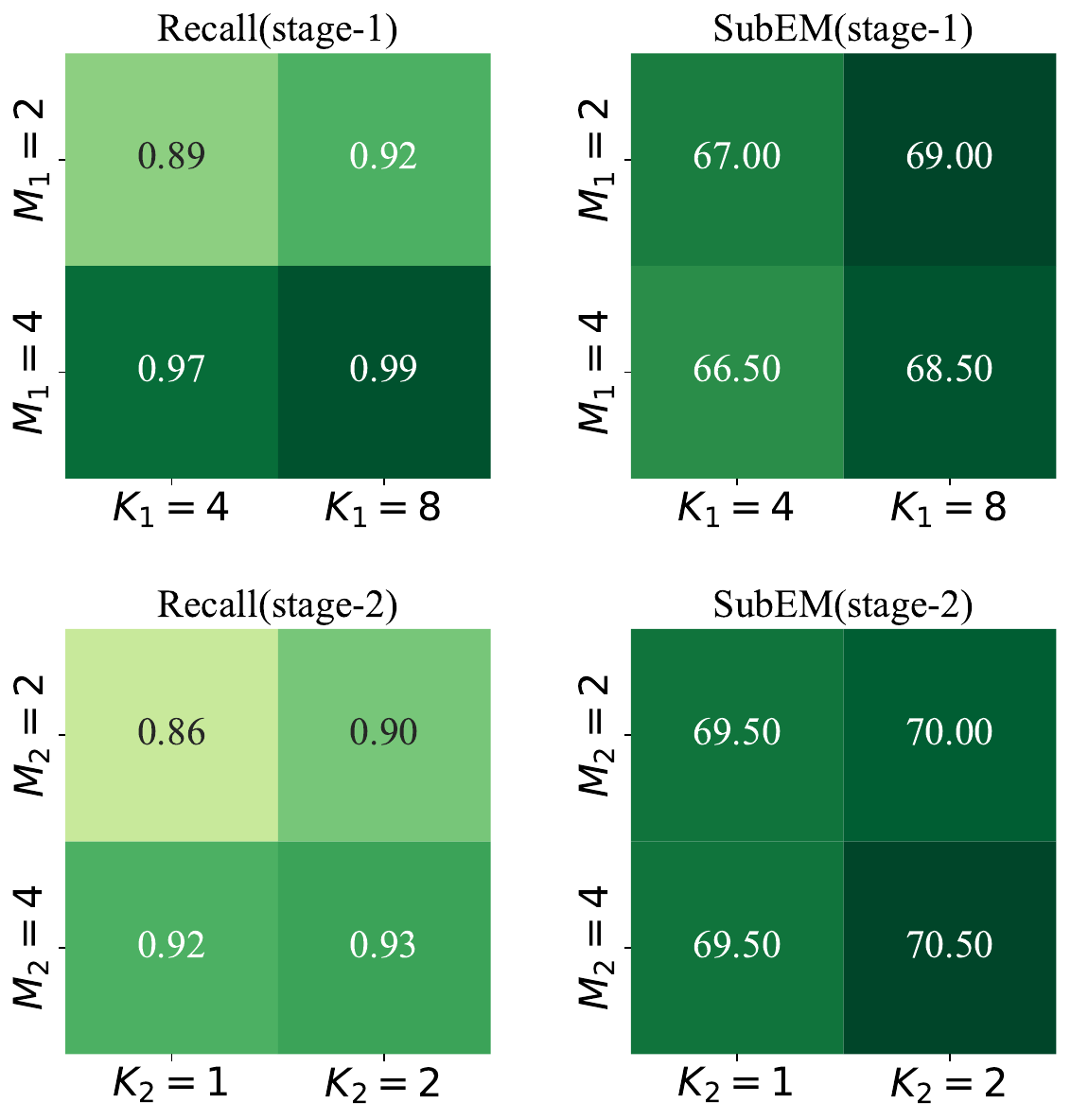}
  \caption{The impact of hyperparameters \(M\) (documents per retrieval head) and \(K\) (retrieval heads) on Llama-3.1-8B-Instruct. The top row shows recall and performance for coarse-grained filtering, while the bottom row illustrates changes for fine-grained steering.}
  \label{fig:hyperparameters-1}
\end{figure}

\paratitle{Impact of Hyperparameters.}
The choice of hyperparameters \(M\) (documents per head) and \(K\) (number of heads) during retrieval head selection has a strong influence on both recall and overall performance. As shown in Figure~\ref{fig:hyperparameters-1}, increasing \(M_1\) or \(K_1\) boosts recall by adding more candidates, but can also introduce noise that limits final accuracy. We therefore fix \(M_1=4\) and \(K_1=4\) for consistency. A similar trade-off holds in the second stage, though performance remains stable after attention steering. The remaining hyperparameter details are provided in Appendix ~\ref{app:hyperparameter}.

\paratitle{Granularity of Attention Steering.}
In the fine-grained steering stage, we also evaluate the impact of granularity of attention steering. Instead of document-level, we identify relevant contexts at sentence-level and steer the attention scores on these sentences. As shown in Table~\ref{tab:hotpotqa_granularity}, the recall at sentence level is lower compared to document level. Additionally, the final performances degrade significantly, even  inferior to that before attention steering. This indicates the importance of covering the golden evidence information as much as possible during the attention steering stages.

\paratitle{Lost-in-the-Middle Performance.}
We investigate the Lost-in-the-Middle phenomenon and the effectiveness of our method in mitigating it. Experiments are conducted on the SQuAD dataset using LLaMA and Mistral, evaluating how the position of the answer within a set of 50 documents affects model performance. As shown in Table~\ref{tab:lost_in_middle}, the Lost-in-the-Middle phenomenon significantly degrades the baseline method's performance, particularly when answers are in middle positions (e.g., Mistral's SubEM score drops from 70\% to 58\%). Our method effectively mitigates the  issue, achieving stable and significantly improved performance across all answer positions, consistently outperforming the baseline. This approach demonstrates strong robustness and generalizability, requiring no position-specific adjustments.


\section{Conclusion}
In this paper, we explored the challenges faced by LLMs in handling long-context inputs, particularly in multi-document question answering tasks. Our findings revealed that the inclusion of irrelevant documents significantly hampers the retrieval and reasoning capabilities of LLMs, motivating the need for more effective long-context processing strategies. To address this, we introduced CAFE, a two-stage coarse-to-fine information-seeking method that leverages retrieval head-based filtering, document reranking, and fine-grained attention steering to guide LLMs in processing long-context inputs. 
Extensive experiments across multiple benchmarks and LLMs validate its effectiveness, demonstrating its superiority over strong baselines, including supervised fine-tuning techniques.
Beyond its performance benefits, CAFE's training-free nature and broad applicability make it a practical solution for a wide range of downstream tasks.

\section*{Limitations}
In this paper, we present a coarse-to-fine two-stage framework to enhance retrieval and reasoning capacities of LLMs. Beyond multi-document question answering tasks, we believe our framework can be employed in broader tasks, \eg long-document reasoning, which have not been explored owing to the computational costs. Additionally, our method mainly focus on how to better identify evidence documents to enhance performances. However, though given the golden evidence, the LLMs can still hardly to answer each question correctly. Approaches of improving the context-aware reasoning capacities can be employed to further improve the  upper limit of our method.

\bibliography{custom}

\appendix

\clearpage
\appendix

\section{Performance Gap}
We study how distracting documents affects model performance in long-context settings. Starting with only gold evidence, we gradually insert irrelevant documents and observe a performance drop as shown in Figure~\ref{fig:exp-gold-other}. This suggests that longer inputs with more irrelevant content weaken the model’s retrieval and reasoning. This motivates our design of a retrieval strategy to filter out such noise.

\section{Baselines}
\label{app:baselines}
We compare CAFE with the following baselines:

\(\bullet\) \textbf{Directly Answering.} Asking LLMs to directly answer the question by using the context.

\(\bullet\) \textbf{In-Context Retrieval.} LLMs are initially prompted to generate the key documents that support answering the question. Then, models are prompted to answer the question with the key documents appended to the context.

\(\bullet\) \textbf{Oracle RAG.} Asking LLMs to answer the question only based on the ground-truth documents to estimate an upper limit performance.

\(\bullet\) \textbf{Vanilla RAG.} For Retrieval-Augmented Generation (RAG) over the documents, we employ BGE-large-env-1.5~\cite{xiao-sigir-2024-cpack} as the embedding model.

\(\bullet\) \textbf{Supervised Fine-tuning.} The LLM is trained on training sets of these datasets. We randomly sample 2000, 5000, and 5000 training instances for SQuAD, HotpotQA, and MusiQue, respectively.

\section{Evidence Selection Results}
\label{app:evidence}
We conduct additional validations on larger and different models, and the experimental results are shown in Table~\ref{tab:combined-retrieval-results}.
Even for Llama-3.1-70B-Instruct, a model with significantly more parameters, its ICR (In-Context Retrieval) capability still declines sharply as the context length increases, whereas the attention-based retrieval method remains more stable. This suggests that ICR is constrained by the expansion capability of the context window, whereas attention-based methods better adapt to long-text settings.
Additionally, we supplement our experiments with Mistral and Phi, among other models, to further validate the generalizability of our findings. The results consistently demonstrate that attention-based retrieval is more robust than ICR in long-context scenarios.

\begin{table*}[htb]
  \centering
  \small
  \setlength{\tabcolsep}{6pt}
  \begin{tabular}{llccccc}
    \toprule
    \textbf{Model} & \textbf{Method} & \textbf{HotpotQA-8k} & \textbf{HotpotQA-16k} & \textbf{HotpotQA-32k} & \textbf{SQuAD} & \textbf{Musique} \\
    \midrule
    \multirow{2}{*}{Llama-3.1-70B-Instruct}
      & ICR
      & 0.82 & 0.72 & 0.54 & 0.92 & 0.67 \\
      & Attention-Based
      & 0.85 & 0.81 & 0.77 & 0.95 & 0.81 \\
    \midrule
    \multirow{2}{*}{Mistral-3-7B-Instruct}
      & ICR
      & 0.65 & 0.49 & 0.33 & 0.78 & 0.46 \\
      & Attention-Based
      & 0.75 & 0.71 & 0.64 & 0.82 & 0.60 \\
    \midrule
    \multirow{2}{*}{Phi-3.5-Mini-Instruct}
      & ICR
      & 0.39 & 0.27 & 0.21 & 0.72 & 0.33 \\
      & Attention-Based
      & 0.54 & 0.52 & 0.43 & 0.73 & 0.36 \\
    \bottomrule
  \end{tabular}
  \caption{
    SubEM scores comparing In-Context Retrieval (ICR) and Attention-Based Retrieval methods across different context lengths and datasets. 
    The second column indicates the retrieval method; performance declines for ICR as context length grows, while the attention-based approach remains more robust.
  }
  \label{tab:combined-retrieval-results}
\end{table*}

\begin{table*}[htb]
  \centering
  \small
  \begin{tabular}{lccccc}
    \toprule
    \textbf{Method} & \textbf{HotpotQA-8k} & \textbf{HotpotQA-16k} & \textbf{HotpotQA-32k} & \textbf{SQuAD} & \textbf{Musique} \\
    \midrule
    Directly Answering & 637.02 & 1482.25 & 2867.17 & 592.65 & 1493.73 \\
    Ours (prefill)     & 840.95 & 1799.92 & 3880.35 & 719.55 & 1791.15 \\
    \bottomrule
  \end{tabular}
  \caption{TTFT (ms/token) comparison across datasets. “Ours (prefill)” refers to the inference time including the prefill enhancement.}
  \label{tab:ttft-results}
\end{table*}

\section{Differences Between Coarse and Fine Retrieval Heads}
\label{app:diff_retrieval_heads}
\begin{figure}[htb]
  \centering
  \includegraphics[width=\columnwidth]{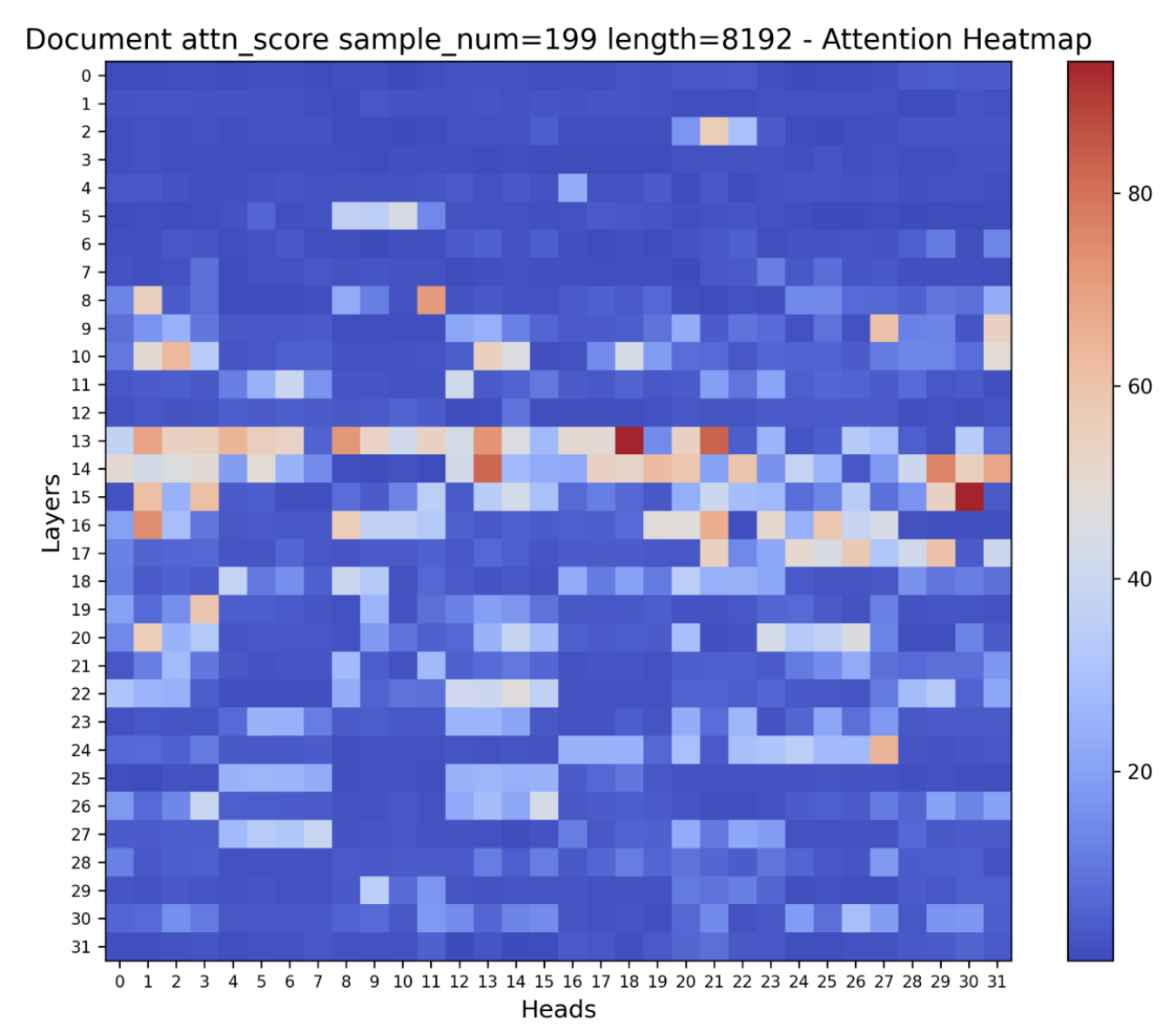}
  \caption{Attention distribution of retrieval heads used in the coarse-grained filtering stage.}
  \label{fig:coarse-heatmap}
\end{figure}

\begin{figure}[htb]
  \centering
  \includegraphics[width=\columnwidth]{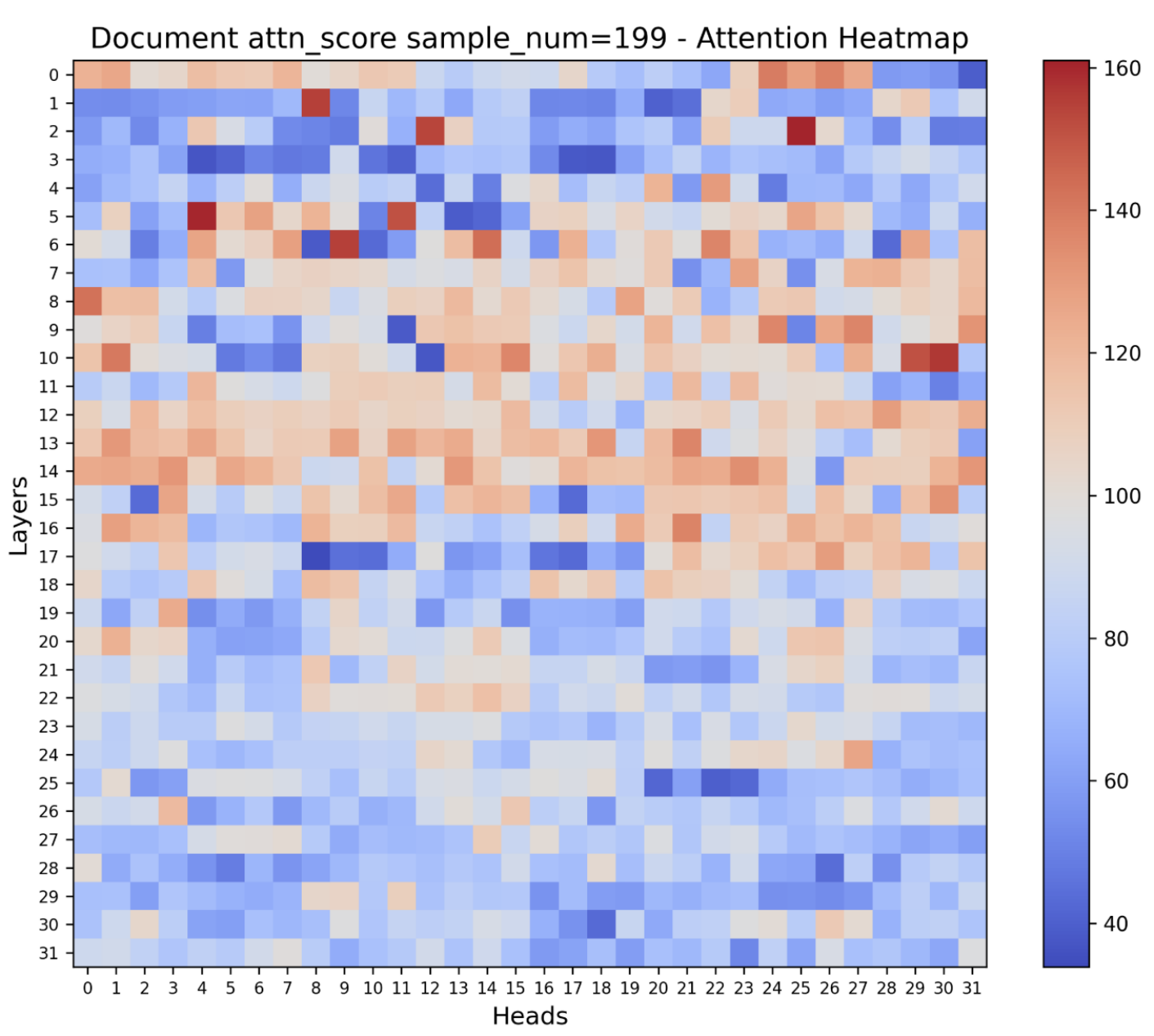}
  \caption{Attention distribution of retrieval heads used in the fine-grained steering stage.}
  \label{fig:fine-heatmap}
\end{figure}

To better understand the behavior of retrieval heads used in the two stages, we visualize their accumulated attention scores over documents in Figure~\ref{fig:coarse-heatmap} and Figure~\ref{fig:fine-heatmap}. We observe that coarse-stage retrieval heads focus on a few documents to filter background information, while fine-stage heads attend more broadly to distinguish gold evidence from distractors.

\section{Impact of Hyperparameters}
\label{app:hyperparameter}
We perform an ablation study on the fine-grained retrieval parameter $\delta$. As shown in Table~\ref{tab:delta-ablation}, values around $\log 10$ yield stable performance, and we use $\log 13$ as the default setting in our main experiments.

\begin{table}[htb]
  \centering
  \small
  \begin{tabular}{lcccc}
    \toprule
    $\delta$ & $\log 5$ & $\log 10$ & $\log 13$ & $\log 20$ \\
    \midrule
    EM score & 69.0 & \textbf{70.5} & 70.0 & 69.5 \\
    \bottomrule
  \end{tabular}
  \caption{Ablation study on the fine-grained retrieval parameter $\delta$.}
  \label{tab:delta-ablation}
\end{table}

\section{Inference Latency Evaluation }
\label{app:Latency}
Our method requires multiple inferences (at least two prefilling operations), which indeed increases inference latency.
In the first round, the prefill length is the same as that of direct answer. As a result, our method is slightly slower than the native flash-attention used in direct answer. We report the difference in prefill efficiency between our method and the Directly Answering baseline in Table~\ref{tab:ttft-results}


\end{document}